\documentclass{article}
\usepackage{amssymb}
\usepackage[utf8]{inputenc} % allow utf-8 input
\usepackage[T1]{fontenc}    % use 8-bit T1 fonts
\usepackage[backref=page]{hyperref}
\usepackage{url}            % simple URL typesetting
\usepackage{booktabs}       % professional-quality tables
\usepackage{amsfonts}       % blackboard math symbols
\usepackage{nicefrac}       % compact symbols for 1/2, etc.
\usepackage{microtype}      % microtypography
\usepackage{xcolor}         % colors
\usepackage{wrapfig}
\usepackage{multirow}
\usepackage{amsmath}
\usepackage{amssymb}
\usepackage{cleveref}
\usepackage{wrapfig}
\usepackage{enumitem}
\setlist{leftmargin=6mm}
\usepackage{graphicx} 
\usepackage{array} 
 \usepackage{algorithm}
\usepackage{algorithmic}
\usepackage{wrapfig}
\usepackage{authblk}

\usepackage{subcaption}
\usepackage{titlesec}
\usepackage{diagbox}

\newcommand{\ours}{DexMachina}
\usepackage[preprint]{corl_2025} % Uncomment for pre-prints (e.g., arxiv); This is like ``final'', but will remove the CORL footnote.

\title{DexMachina: Functional Retargeting for\\ Bimanual Dexterous Manipulation}

\author{
    \textbf{Zhao Mandi$^{1\ *}$, Yifan Hou$^{1}$, Dieter Fox$^{2}$,
    Yashraj Narang$^{2}$, Ajay Mandlekar$^{2,\dag}$, Shuran Song$^{1,\dag}$} \\ \\
    $^{1}$Stanford University \quad $^{2}$NVIDIA \quad $^{\dag}$equal advising \quad $^{*}$work partially done at internship
}

\begin{document}
\maketitle

%===============================================================================
% motivating object controller:
% Compare with chunking

\begin{abstract}
% We study the problem of functional retargeting from human hand demonstrations to bimanual dexterous manipulation policies. Direct learning is difficult in this setting due to sparse exploration and embodiment gap between human and robot hands. We propose a novel algorithm, \ours, which uses a curriculum that first guides the policy through the entire motion, then let the policy gradually take over. We build a simulation benchmark with a diverse set of dexterous robot hand designs, and show that \ours~achieves functional retargeting on a variety of hands, articulated objects, and long-horizon tasks. Having this effective algorithm now unlocks a new framework for functional evaluation and comparison of different hardware designs. With the recent surge in dexterous hand development, we hope our algorithm and benchmark environments will provide a useful platform for identifying desirable hardware capabilities and will lower the barrier for contributing to future research. 
We study the problem of functional retargeting: learning dexterous manipulation policies to track object states from human hand-object demonstrations. We focus on long-horizon, bimanual tasks with articulated objects, which is challenging due to large action space, spatiotemporal discontinuities, and embodiment gap between human and robot hands. We propose DexMachina, a novel curriculum-based algorithm: the key idea is to use virtual object controllers with decaying strength: an object is first driven automatically towards its target states, such that the policy can gradually learn to take over under motion and contact guidance. We release a simulation benchmark with a diverse set of tasks and dexterous hands, and show that DexMachina significantly outperforms baseline methods. Our algorithm and benchmark enable a functional comparison for hardware designs, and we present key findings informed by quantitative and qualitative results. With the recent surge in dexterous hand development, we hope this work will provide a useful platform for identifying desirable hardware capabilities and lower the barrier for contributing to future research. Videos and more at \url{project-dexmachina.github.io}
\end{abstract}

 % \begin{figure}[hbt!]
 %    \centering    \includegraphics[width=0.98\linewidth]{figs/dexmachina-teaser-0425v1.pdf}
 %    \caption{\small We study the problem of functional retargeting, where the goal is to retarget human hand demonstrations into functional dexterous robot policies that manipulate an object to follow the desired trajectory. Our proposed algorithm achieves retargeting from one human demonstration to a variety of many existing dexterous hand hardwares. 
 %    % shuran: show object movement in one view. show only the  hand in first row. 
 %    }
    %\vspace{-3mm}
% \end{figure}
 \begin{figure}[hbt!]
    \centering    \includegraphics[width=0.98\linewidth]{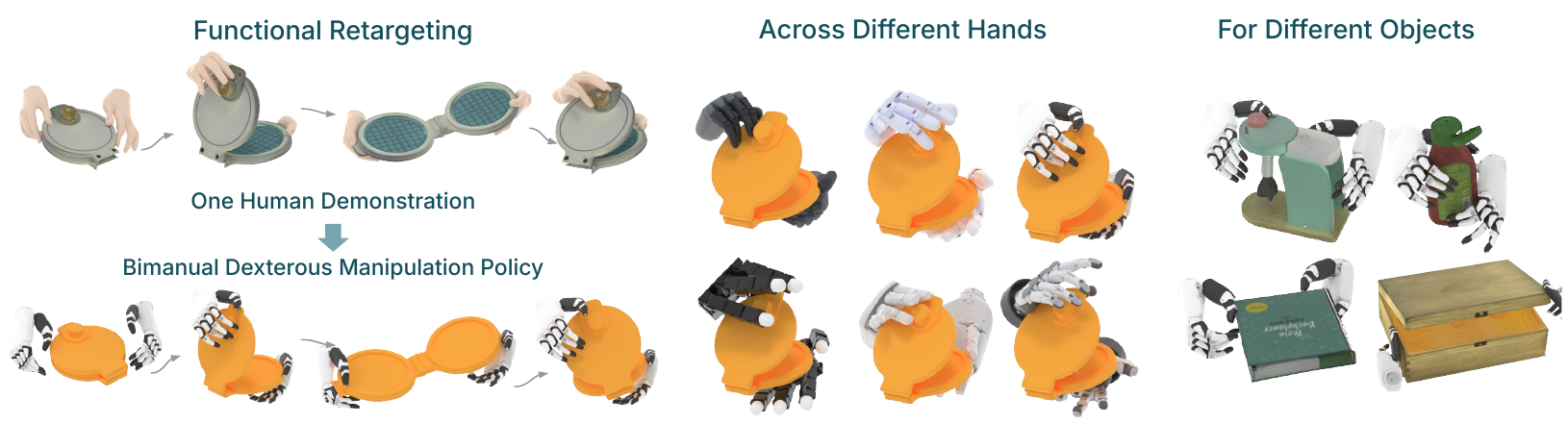}
    \caption{\small \textbf{Functional Retargeting.} We study the problem of functional retargeting, where the goal is to retarget human hand demonstrations into functional dexterous robot policies that manipulate an object to follow the demonstrated trajectory. Our proposed algorithm, \ours, achieves functional retargeting from one human demonstration to a variety of existing dexterous hand embodiments over a range of articulated objects. 
    % Anonymous submission website: \url{dexmachina-submission.github.io} 
    }
    \label{fig:teaser}
    %\vspace{-4mm}
\end{figure}
%===============================================================================
% Outline: (0328) 
% 1. We work on "funtional retargeting", how it is different and better than kinematic retargeting: allows human hand to many robot hand.  
% 2. functional retargeting is hard and require new method
% 3. our embodiment agnostic algorithm ... using auto-curriculum.  
% 4. With our algorithm, it is now possible to benchmark different hand designs for a variety of tasks (in the past, making one hand work for one task was hard already). Key finding: something surprising, similarity to human hand size is less important than total DOF. 
\vspace{-4mm}
\section{Introduction}\label{sec:intro} 
\vspace{-4mm}
% Dexterous manipulation promises new task capabilities in both industrial and household settings, but has seen limited progress due to bottlenecks in both hardware and algorithms. Due to the complexity of dexterous hand motion, a popular approach is kinematic retargetting~\cite{qin2023anyteleop}: given human hand demonstrations, compute robot hand motion that looks like the human. In this work, we study the problem that we denote as \textit{functional retargeting}, where we additionally require the learned motion to be not only similar to the demonstration but also dynamically feasible to achieve the demonstrated manipulation task.  In this work, we aim to push the limits of algorithmic performance on generalization and robustness across different dexterous hands. 
% The past decades have seen continuous efforts in developing new hardware designs and sensing capabilities~\cite{Belgrade,stanford-jplhand, utah-mithand,dlr-hand1, shiokata2005robot, higo2018rubik}, however, human-like general purpose dexterity is still a distant goal. We argue the bottleneck lies in algorithms: due to a lack of functioning algorithms and standardized benchmarks for multiple hand designs, researchers do not fully understand the hardware requirements for solving complex manipulation tasks, and hence continue to build varying dexterous hands independently and develop methods that are validated on specific platforms. 
Dexterous robot hands, with their resemblance to human hands, spark the expectation for achieving human-level dexterity. Yet the reality presents many hardware and algorithmic challenges that bottleneck the progress in dexterous manipulation. Prior learning-based methods have seen success in relatively simple and short-horizon tasks, but are often limited by manual reward-engineering~\cite{openaihand,DextrAH-G} or costly data-collection~\cite{qin2023anyteleop,openaihand} due to the embodiment gap between human hands and dexterous hands. 
% Dexterous robot hands, with their resemblance to human hands, raises the enormous promise in achieving human-level manipulation dexterity. 
% However, there are several hardware and algorithmic challenges that present significant bottlenecks for achieving such dexterity. 
% Dexterous robot hands, with their resemblance to human hands, spark the expectation that they should learn from human hand dexterity and unlock new task capabilities. 
% However, the reality presents many hardware and algorithmic challenges that bottleneck the progress in dexterous manipulation. 
% Despite continuous effort in developing new hands and sensing capabilities~\cite{Belgrade,stanford-jplhand, utah-mithand,dlr-hand1, shiokata2005robot, higo2018rubik}, there lacks effective learning algorithms for and standardized evaluation benchmark

Human hands are hence a natural source for learning guidance. In this work, we formulate learning from human with an emphasis on task capability. We denote the problem as \textit{functional retargeting}: given a human demonstration, the goal is to learn dexterous hand policies that can manipulate the object to follow the demonstrated trajectory (see Fig.~\ref{fig:teaser}). This is distinguished from \textit{kinematic} retargeting~\cite{qin2023anyteleop}, which produces human-like motions without ensuring feasibility. The problem is even more compelling for long-horizon, bimanual demonstrations with articulated objects, which encompass a significant portion of daily human activities, but pose several key challenges: exploration is difficult under the high-dimensional action space, the intricate contact sequences demand stable and precise hand movements; due to the embodiment gap, human hand motion cannot be directly mapped to feasible robot actions, which limits the scalability of imitation data collection.

To address these challenges, we propose~\ours\footnote{Deus ex machina (``god from the machine"), is when a seemingly unsolvable problem is conveniently solved by an external force --- much like how our algorithm moves an object by itself before the policy gradually learns to take over, hence the name \ours.}, a novel curriculum-based RL algorithm for functional retargeting. Precise bimanual coordination is often required to manipulate an object successfully (e.g. opening a waffle iron mid-air, see Fig.~\ref{fig:teaser}), but naive approaches often get stuck in early failures or suboptimal actions. This motivates us to design a curriculum to allow the policy to explore in a less fragile setting. Our key idea is to use \textit{virtual object controllers}---they apply control forces that drive an object towards its demonstrated trajectory---and \textit{auxiliary motion and contact rewards}, which guide the policy to learn task strategies as the virtual controller strength decays. The policy first learns to mimic the human motion without worrying about failing the task, then learns to take over manipulation as the virtual controllers fade away. 
Despite continuous effort in developing new hands and sensing capabilities~\cite{Belgrade,stanford-jplhand, utah-mithand,dlr-hand1, shiokata2005robot, higo2018rubik}, there is a lack for standardized and accessible evaluation benchmarks. To address this, we build a simulation benchmark with a diverse set of 6 dexterous hands and 5 articulated objects~\cite{fan2023arctic}, and provides a unified testbed where new hands and tasks can be easily added and quickly evaluated. On this benchmark, we empirically show that \ours~significantly outperforms baseline methods, and applies successfully to a wide variety of hands, articulated objects, and long-horizon demonstrations. 
% Leveraging massively-parallelized physics simulation~\cite{Genesis}, we are able to quickly train dexterous policies within hours for a new hand or a new demonstration. 
% how suitable a given hand design is for learning from human motion.
% \shuran{We propose a new way to evaluate the hand hardware capability, i.e., "use learned policy performance as a way to benchmark the hardware capability." % Then you can justify why it is a good approximation: 1. why it is meaningful? it measures both functionality and easiness to learn from human demonstration 2. why it is fair? i.e., why is your algorithm not biased towards some hand? It is hand agonistic? 3. why is it practical? With massively parallelized simulation, it is easy to compute.  

With an effective algorithm and evaluation benchmark for functional retargeting, it is now possible to make functional comparisons across different hardware: informed by the policy learning performance, we obtain a meaningful measure for both the hands' functionality and readiness to learn from human guidance. This comparison is generalizable and accessible: our algorithm requires no hand-specific adaptations, and our task environments are fast to run and easy to customize. With the recent surge in the development of robotic hand hardware, we hope this functional comparison will be helpful for making informed decisions for both acquiring and designing new hands. 

\textbf{Our contributions are summarized as follows:}
\newline\noindent $\bullet$ We study \textbf{Functional Retargeting}, where we learn feasible dexterous manipulation policies from human hand-object demonstrations. We propose \textbf{\ours}, a novel algorithm for functional retargeting based on a curriculum over virtual object controllers and motion and contact guidance.
\newline\noindent $\bullet$ We introduce the \ours~benchmark with 6 curated dexterous hand assets and 5 articulated objects, for evaluating both different functional retargeting algorithms and robotic hand designs. 
% We assess both learning efficiency under human hand guidance and functional utility from the task performance on a diverse set of objects and motions.
\newline\noindent $\bullet$ We demonstrate \ours~achieves state-of-the-art learning performance across a variety of robotic hands and tasks. Our simulation environments and learning algorithms will be open-sourced to facilitate future research. 
% To summarize our contributions: we study the problem of functional retargeting, where we learn feasible dexterous manipulation policies from human hand-object demonstrations. We propose \ours~, a novel algorithm for functional retargeting based on a curriculum over virtual object controllers and motion and contact guidance. We demonstrate state-of-the-art learning performance across a variety of robotic hands. We \ourbench, a benchmark that uses our algorithm as a framework for evaluating robotic hand designs, where we assess both learning efficiency under human hand guidance, and functional utility from the task performance on a diverse set of objects and motions. Our simulation environments and learning algorithms will be open-sourced to facilitate future research. 

% Given the recent surge in the development of new robotic hand hardware~\cite{inspire,xhand}, our algorithm and benchmark enable principled evaluation of a hand's capability to perform human-like manipulation skills based on how well its policy achieves the functional goal, thus helping researchers and engineers make decisions and trade-offs on hand design

% Shuran 3.28 comments
% 1. the task of functional Retargeting --  human hand to many robot hand
% 2. the embodiment-agnostic method using curriculum 
% 3. the benchmark that evaluates the hand designs and the key findings 

%=============================================================================== 
\vspace{-3mm}
\section{Related Work}
\label{sec:related} 
\vspace{-3mm}
\textbf{Reinforcement Learning for Dexterous Manipulation.} 
Reinforcement Learning (RL) has been used for dexterous manipulation tasks such as in-hand object orientation~\cite{openaihand, handa2023dextreme, qi2023hand,yin2023rotating,chen2023visual} and single-hand grasping~\cite{DextrAH-G,Caggiano2023MyoDexAG,Luo2024GraspingDO,mandikal2021graff, Zhu2023TowardHG,Yuan2024CrossDex}, but achieving more complex, longer-horizon manipulation remains challenging due to the burden of designing rewards to guide exploration for such tasks.
Model-based methods have been applied to tasks such as ball dribbling~\cite{shiokata2005robot} and Rubik's cube turning~\cite{higo2018rubik}, but they require careful engineering for each object and task.
In our work, we seek to study bimanual long-horizon tasks where it can be difficult to specify concrete goals or design RL rewards to guide exploration. This motivates our use of human demonstrations, which both act as a goal specification and provide guidance for how to solve the task.
Simulation is a common tool to train dexterous hand policies~\cite{DAPG} due to the high exploration cost of running RL on real hardware~\cite{Xu2022DexterousMF}.
Our simulation benchmark supports evaluation across several dexterous hands and diverse tasks defined by human demonstration data, in contrast to existing RL benchmarks~\cite{dexart,shadow} for dexterous manipulation.

\textbf{Imitation Learning for Dexterous Manipulation.} Imitation learning (IL) is a compelling alternative to RL, since the the use of demonstrations can mitigate or eliminate the burden of exploration, but it can require accurate on-robot action data that is challenging to capture for dexterous hands. Most existing approaches ~\cite{qin2023anyteleop, wang2024dexcap, yang2024ace, opentelev, shaw2024bimanual, zhang2025doglove} require setting up a teleoperation system customized for a particular robot hand embodiment.
Human hand data (such as videos) are another source of data. 
Prior work has used human hand data for learning rough grasp affordances~\cite{Mandikal2020LearningDG}, improved retargeting~\cite{park2025learningtransferhumanhand}, or co-training with human hand data and teleoperation data~\cite{videodex,xu2023xskill}, but these approaches have been limited for short-horizon manipulation (mainly grasping). 
Instead, our work assumes access to a single tracked hand-object demonstration per task and uses the demonstration to guide RL training. Similar approaches have been used for humanoid locomotion~\cite{deepMimic}, simple hand manipulation~\cite{physicsHOI}, and dexterous manipulation on short-horizon tasks~\cite{Chen2024ObjectCentricDM, li2025maniptrans}.

% \paragraph{Learning from Dexterous Demonstration}
% Imitation learning (IL) is alternative approach to RL, but it requires accurate action data that is challenging to capture for dexterous hands. Most existing approaches ~\cite{qin2023anyteleop, wang2024dexcap, yang2024ace, opentelev, shaw2024bimanual, zhang2025doglove} require setting up a teleoperation system customized for a particular robot hand embodiment. 

% % Generalization to multiple embodiments was achieved for grasping by only using the teleoperation data to compute grasp patterns, then applying RL~\cite{Yuan2024CrossDex}.
% %\vspace{-2mm}
% \paragraph{Learning Dexterous Manipulation from Human Hands}
% Unstructured human-hand data, such as videos, although they cannot directly produce accurate robot actions due to the embodiment gap, prior work has used human hands for learning rough grasp affordances~\cite{Mandikal2020LearningDG} or co-trained with teleoperation data~\cite{videodex}, but they only work for short-horizon, mainly grasping tasks. In contrary, we make a stronger assumption for accurately-tracked human hand-object demonstrations, which is more expensive to collect but provides rich information for task goals and strategies, and requires only one demonstration and leave the rest to RL training. The idea was proven effective on humanoid locomotion~\cite{deepMimic}, simple hand manipulation~\cite{physicsHOI}, and dexterous manipulation on short-horizon tasks~\cite{Chen2024ObjectCentricDM, li2025maniptrans}.
%\vspace{-2mm}

\begin{figure}[t]
    \centering    \includegraphics[width=0.98\linewidth]{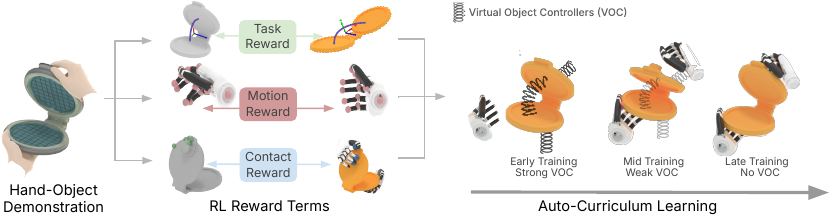}
    \caption{\small \textbf{\ours\ Overview.} \ours~is a curriculum-based RL algorithm for functional retargeting. We process densely-tracked human hand demonstration to extract reference robot joints and keypoints (pink spheres) and approximated contact positions on object mesh vertices (green spheres), which we use to define auxiliary rewards in addition to the task reward. We then introduce an auto-curriculum using virtual object controllers, which initially moves the object on its own to follow the demonstration, and are then decayed over the course of RL training as the policy learns to take over manipulation.
    }
    \label{fig:method}
    % \vspace{-2mm}
\end{figure}

\textbf{Curriculum Learning.}
It is common practice in optimization-based motion planning to warmstart an optimization from relaxed physical constraints, resulting in a better solution at convergence~\cite{mordatch2012discovery,pang2021convex,pang2023global}.
This idea of learning using a curriculum, which moves from easier to more difficult problems, has been adopted by RL methods~\cite{sds-curriculum, zhang2024artigrasp}. Some prior work uses this approach to relax physical constraints, such as allowing force before making contact~\cite{mao2025learning} or relaxing gravity, friction and constraint-solver parameters~\cite{li2025maniptrans}.
Our approach uses a curriculum over object dynamics, allowing the agent to gradually learn how to manipulate the object over time (Fig.~\ref{fig:method}).

\vspace{-3mm}
\section{Functional Retargeting Formulation} 
\vspace{-3mm}
We define the \textit{functional retargeting} problem as follows: given one object $\eta$, one human hand-object demonstration sequence $\mathcal{D}^{\eta}$, and a pair of dexterous robot hands $\zeta$, the goal is to learn a robot policy that can manipulate the object to track the demonstrated object states. More formally, one human demonstration $\mathcal{D}^{\eta}=\{G, H\}$ contains $T$ timesteps of densely tracked object states $G$ and hand poses $H$. We focus on articulated objects, hence the object states include both part pose and revolute joint angle values. At any timestep $t$, given an achieved object state $\hat{g_t}$ (position, rotation, and articulation) and the target object state from the demonstration $g_t = \{g_t^P,g_t^R, g_t^J\}$, we denote the distance function as $F$ (computes both rotation, position, and articulation joint error). The learned policy for $(\eta, \zeta)$ should minimize the accumulated tracking error across all timesteps: $\pi_{\theta}^{\eta, \zeta}= \text{argmin}_{\theta} \sum_{t=1}^T(F(\hat{g_t},g_t))$.
% NOTE will add below back for arxiv, saving space for now
% \begin{equation}
%     \pi_{\theta}^{\eta, \zeta}= \text{argmin}_{\theta} \sum_{t=1}^T(F(\hat{g_t},g_t))
% \end{equation}\label{eq:track}

%\vspace{-10mm} 
% , i.e. $G\in \mathbb{R}^{T\times8}$.  
% We use bimanual hand demonstrations represented by MANO~\cite{MANO:SIGGRAPHASIA:2017}, which tracks 6-DoF wrist poses and 21 fingertip keypoint positions, hence $H = \{H_{\text{left}}, H_{\text{right}}\}, H^{\text{left}} = \{ H_{\text{left}}^{\text{wrist}} \in\mathbb{R}^{T\times6}, H_{\text{left}}^{\text{fingertip}} \in\mathbb{R}^{T\times21\times3} \} 

\vspace{-3mm}
\section{Method}
\vspace{-3mm}
% \mandi{new outline:
% 1. first describe task reward, object pose tracking etc. 2. kinematic retargeting, data processing. 3. use retargeted results to get base actions; 4. imitation and contact rewards. 5. curriculum 
% } 
\textbf{Overview.}
 We propose \ours, a curriculum-based RL algorithm for functional retargeting. In \S\ref{method:task}, we begin by introducing the task reward, which encourages object tracking in but is insufficient for effective policy learning. In \S\ref{method:aux}, we extract motion and contact information from demonstrations, which we use to define residual actions and auxiliary rewards. While these components improve learning, they still fall short in complex long-horizon tasks. This motivates our curriculum strategy, presented in \S\ref{method:curr}, where we introduce an auto-curriculum based on virtual object controllers to achieve efficient functional retargeting across different dexterous hands.

% \vspace{-2mm}
\subsection{RL Environment and Task Reward}\label{method:task}
%\vspace{-3mm} 
We train reinforcement learning (RL) policy to achieve the functional retargeting task. An RL environment is constructed by pairing one demonstration $\mathcal{D}^{\eta}$ and one set of bimanual dexterous robot hands $\zeta$. At each timestep $t$, write $G_t=\{g_t^P, g_t^R,g_t^J\}$ for the recorded object position, rotation and joint angles at timestep $t$, and $\hat{G_t}=\{\hat{g}_t^P, \hat{g}_t^R, \hat{g}_t^J\}$ for the object's achieved states corresponding to each term. The task reward $r_{\text{task}}$ is the product of three terms measuring accuracy in each state component, encouraging balanced learning~\cite{Chen2024ObjectCentricDM}. Formally: 
$$d_{\text{pos}} = ||\hat{g_t}^{T}-g_t^T||_2; \ d_{\text{rot}} = 2 \cos^{-1}(|\langle \hat{g_t}^R, g_t^R\rangle|) ; \ d_{\text{ang}} = ||\hat{g_t}^{J}-g_t^J||_2$$
$$
r_{\text{task}} = r_{\text{pos}} * r_{\text{rot}} * r_{\text{angle}} = \exp(-\beta_{\text{pos}} d_{\text{pos}}) \exp(-\beta_{\text{rot}} d_{\text{rot}}) \exp(-\beta_{\text{ang}} d_{\text{ang}}) 
$$

% $$r_{\text{task}} = r_{\text{pos}} * r_{\text{rot}} * r_{\text{angle}} = \exp(-\beta_{\text{pos}} ||\hat{g_t}^{T}-g_t^T||_2 - \beta_{\text{rot}} d_{\text{rot}} -\beta_{\text{ang}} d_{\text{ang}})
% $$
% $$r_{\text{task}} = r_{\text{pos}} * r_{\text{rot}} * r_{\text{angle}} = \exp(-\beta_{\text{pos}} ||\hat{g_t}^{P}-g_t^P||_2  - \beta_{\text{rot}} 2 \cos^{-1}(|\langle \hat{g_t}^R, g_t^R\rangle|) -\beta_{\text{ang}} ||\hat{g_t}^{J}-g_t^J||_2)
% $$
% $$r_{\text{task}} = \exp(-\beta_{\text{pos}} ||\hat{g_t}^{P}-g_t^P||_2  - \beta_{\text{rot}} 2 \cos^{-1}(|\langle \hat{g_t}^R, g_t^R\rangle|) -\beta_{\text{ang}} ||\hat{g_t}^{J}-g_t^J||_2)
% $$
% \( \hat{g}_t^P \) and \( g_t^P \) represent the current and demonstrated positions. \( \hat{g}_t^R \) and \( g_t^R \) are unit quaternions for the current and demonstrated rotations. \( \hat{g}_t^J \) and \( g_t^J \) denote the current and demonstrated joint angles.
where \( \beta_{\text{pos}} \), \( \beta_{\text{rot}} \), and \( \beta_{\text{ang}} \) are scalar weights that control the desirable error scale for each component.
%\vspace{-2mm}
\subsection{Action Formulation and Auxiliary Rewards}\label{method:aux}
%\vspace{-2mm}
% \subsection{Demonstration Data Pre-processing} \label{method:preproc}
Although task reward specifies desired object states, it does not provide useful information for \textit{how} to achieve them. To address this, we (1) propose a hybrid action formulation, which constrains the wrist action space to align more with the human demonstrators; and (2) define auxiliary rewards, which guide the policy to follow the human's hand-object interaction strategy. As a preliminary, we first apply pre-processing on the demonstration data $\mathcal{D}^{\eta}$ to extract relevant motion and contact information. 

\textbf{Data Pre-Processing.}
Given $\mathcal{D}^{\eta}$ with $T$ timesteps, $N$ object parts, and a dexterous hand $\zeta$ with $J$ actuated joints and $K$ collision links, we first run a kinematics-only retargeting algorithm~\cite{qin2023anyteleop} that matches dexterous hand poses with human hand motion. Then we obtain: 
%\vspace{-2mm}
\begin{enumerate} 
\item \textbf{Collision-aware kinematic retargeted joints} $\mathcal{Q}\in\mathbb{R}^{T\times J}$ and \textbf{reference keypoints} $\mathcal{X}\in\mathbb{R}^{T\times K\times3}$  by replaying the retargeting results in simulation and recording (1) the achieved joint values and (2) 3D keypoint positions of the dexterous hand links. To eliminate object penetrations, we replay the retargeted joint values as soft control targets in simulation while keeping the object fixed --- See Appendix~\ref{app:para-ret} for more details.

% Write $x_{t}^k$ for $k$th hand link's position at timestep $t$. 

\item \textbf{Approximated hand-object contact}. Although kinematic retargeting produces human-like dexterous hand poses, the motions often fail to manipulate the object. Hence we extract contact information as additional guidance for object interaction. We use a distance-based approximation to obtain exactly when and where a specific dexterous hand link should be in contact with a specific object part (detailed in Appendix~\ref{app:contact}). The results are approximated contact positions $C\in \mathbb{R}^{(T\times N\times K\times 3)}$ and a mask $M \in \mathbb{R}^{(T\times N\times K)}$ that indicates whether a pair of object part and hand link has valid contacts.
% \ajay{Can we have an additional sentence that describes exactly what these two tensors are? Or maybe link it to a specific example to ground this?}
% by first filtering vertex distances between MANO hand and object meshes, and then grouping to their closest link on the dexterous hands. Each object part is compared to all robot hand links at each timestep. 

% \ajay{Is this implementation new or something standard that we can cite? Also the description is a little vague and confusing right now -- could be worth a little more detail or context. For example we could break it down into a few more steps. Motivate why we want contact information (e.g. to use for reward) and that we want to detect when a specific hand link is in contact with a specific object part. To do this we use a distance-based approximation - etc.}

\end{enumerate}
% ARCTIC articulated object assets and MANO~\cite{MANO:SIGGRAPHASIA:2017} data all have $M=21$ and $N=2$. 
\textbf{Hybrid Action Outputs.} Given the retargeted joint results $\mathcal{Q}$, we use the joint values for 6-DoF wrist joints as base actions, which are added to the policy's output residual actions. The remaining finger joints use absolute actions that are normalized by their joint limits (see Appendix~\ref{app:hybrid} for full details). This formulation effectively constrains the policy's action space, and we empirically find it to significantly improve the learning efficiency. 

% It is also similar to ObjDex~\cite{Chen2024ObjectCentricDM} except their base actions come from a high-level planner trained on human wrist motions in the demonstration data. \ajay{Consider omitting this comparison and mentioning this in related work instead, or appendix section comparing our method vs. theirs explicitly.}

% At timestep $t$, the RL policy outputs actions for all joints:
% $$a_t=\pi_{\theta}(o_t)\in\mathbb{R}^{J} ; \ \hat{q_t}^{(i)} = \gamma^{(i)}a_t^{(i)} + m * q_t^{(i)}$$ 
% Here $o_t$ is the observations, $q_t^{(i)}$ is the retargeted joint results for the $i$th joint, $\hat{q_t}^{(i)}$ is the joint target values sent to the policy's controller, $\gamma^{(i)}$ is the scale. For a joint $i$, $\gamma^{(i)}$ is 10-centimeter if it's a wrist translation joint, 1-radian if it's wrist rotation joint, and 
% While the task reward and constrained action space can guide the policy towards desired object states, they don't specify \textit{how} to achieve it. 
% To address this, we use motion and contact data from \S\ref{method:preproc} to define the following auxiliary rewards that guide the policy to follow the human demonstrator's strategy:
 \textbf{Motion Imitation Reward.} To encourage human-like hand motions, we take the motion reference keypoints $\mathcal{K}$ and retargeted joint values $\mathcal{Q}$, and define (1) motion imitation reward $r_{\text{imi}}$ based on keypoint matching, (2) behavior-cloning reward $r_{\text{bc}}$ based on joint angle distances to the reference. Formally:
    $$\ r_{\text{imi}} = \frac{1}{K} \sum_{i=1}^{K}\exp({-\beta_{\text{imi}} ||\hat{x_i} - x_i||_2}) ; \ r_{\text{bc}} = \frac{1}{J} \sum_{i=1}^{J} \exp({-\beta_{\text{bc}} ||\hat{q_i} - q_i||_2}); \  $$ where each $(\hat{x_i}, x_i)$ denotes the achieved and reference positions for the $i$th keypoint and $(\hat{q_i}, q_i)$ denotes the achieved and retargeted values for the $i$th joint. 

\textbf{Contact Reward.} We read contact positions between each hand link and each object part, and compute contact reward by matching the policy contacts with the corresponding demonstration contacts. For each side of the hand, we denote the policy's and demonstration's contact positions and validity masks as $C, \hat{C}\in \mathbb{R}^{N\times K\times3}$, $M,\hat{M}\in\mathbb{R}^{N\times K\times1}$, respectively. We compute $L_2$ contact distance masked by validity masks and use it to define contact reward $r_{\text{con}}$: 
\begin{equation}
D = \| C - \hat{C} \|_2 \in \mathbb{R}^{N\times K} ;  \ \text{set} \ D^{(i,j)} = 
\begin{cases} 
d_{\text{max}}, & \text{if } M^{(i,j)}_{\text{demo}} \neq M^{(i,j)}_{\text{policy}} \\
0, & \text{if } M^{(i,j)}_{\text{demo}} = M^{(i,j)}_{\text{policy}} = 0
\end{cases} 
\end{equation}\label{eq:contact1}
%\vspace{-3mm}
\begin{equation}
r_{\text{con}} = \frac{1}{2NK} (
\sum_{i=1}^{N}\sum_{j=1}^{K}\exp(-\beta_{\text{con}}D^{(i,j)}_{\text{left}}) + 
\sum_{i=1}^{N}\sum_{j=1}^{K}\exp(-\beta_{\text{con}}D^{(i,j)}_{\text{right}}) 
)
\end{equation}\label{eq:contact2}

% \ajay{Something is messed up above -- you subtract two matrices and take some matrix norm, and then repeat the definition per-entry, and also have an undefined dmax term} 

    % We treat contact positions as a set of point clouds and use Chamfer Distance to measure the distance between demonstrated and observed contacts. A key detail here is to compute the point-cloud distances in both the object root frame and the corresponding hand's wrist frame: this encourages contact at specific locations on the object despite the dexterous robot hands' varying sizes. Contacts are labeled based on the object part ID, and contact rewards are computed separately to ensure the policy makes contact with the correct part. 
    % $$r_{\text{con}} =  \exp(-\beta_{\text{con}}\text{CD}(c_i, c_i^*)) $$

The final RL reward is a weighted sum of the above terms: $r_t = \lambda_{\text{task}}r_{\text{task}} + \lambda_{\text{imi}}r_{\text{imi}} + \lambda_{\text{bc}}r_{\text{bc}} + \lambda_{\text{con}}r_{\text{con}}$. See Appendix~\ref{app:contact} for precise weights and additional reward details.

% \ajay{It might be useful to have a concrete example discussed here to ground what this stuff does, since the discussion in this section is a little abstract and could be hard to follow without considering a specific object and motion.}
\begin{wrapfigure}{R}{0.5\linewidth}
\vspace{-25pt}
\begin{minipage}{\linewidth}
\begin{algorithm}[H]
\small
\caption{\ours~Curriculum}\label{algo:curr}
\begin{algorithmic}[1]  
\REQUIRE Reward thresholds $\sigma_{\text{task}}, \sigma_{\text{imi}}, \sigma_{\text{bc}}, \sigma_{\text{con}}$
\REQUIRE Reward deques $D_{\text{task}}, D_{\text{imi}}, D_{\text{bc}}, D_{\text{con}}$
\REQUIRE Initial gains $k_p, k_v$, decay ratios $\phi_p, \phi_v$, max episode length $L_{\text{max}}$
\FOR{each PPO iteration}
    \FOR{each environment where episode is done}
        \STATE Get achieved episode length $L$
        \STATE Get cumulative rewards $R_{\text{task}}, R_{\text{imi}}, R_{\text{bc}}, R_{\text{con}}$
        \FOR{each term $z \in \{\text{task}, \text{imi}, \text{bc}, \text{con}\}$}
            \STATE Compute normalized reward: $\bar{r}_z = \frac{R_z}{L_{\text{max}}}$
            \STATE Append $\bar{r}_z$ to deque $D_z$
        \ENDFOR
    \ENDFOR
    \FOR{each reward type $z \in \{\text{task}, \text{imi}, \text{bc}, \text{con}\}$}
        \STATE Compute mean: $\mu_z = \text{mean}(D_z)$
    \ENDFOR 
    \IF{$k_p = 0$}
        \STATE \textbf{continue} // no need to decay
    \ENDIF
    \IF{$\mu_z > \sigma_z \ \forall z \in \{\text{task}, \text{imi}, \text{bc}, \text{con}\}$}
        \STATE // Learning is stable, applying gain decay
        \STATE $k_p \leftarrow k_p \cdot \phi_p$
        \IF{$k_p \leq 0.01$}
            \STATE $k_p \leftarrow 0; \quad k_v \leftarrow 0$
        \ENDIF
        \STATE $k_v \leftarrow k_v \cdot \phi_v$
    \ENDIF
\ENDFOR
\end{algorithmic}
\end{algorithm}
\end{minipage} 
\vspace{-10pt}
\end{wrapfigure}

\subsection{Auto-Curriculum with Virtual Object Controllers}\label{method:curr}
% \mandi{Add an example here: if no curriculum, the object just falls before the policy learns other rewards}

\textbf{Motivation.} The above reward terms and action constraints are sometimes sufficient short and simple tasks, but struggle on long-horizon clips with complex contacts. The policy often experiences catastrophic early-stage failures: e.g. after lifting a box with both hands, it might fail to anticipate that one hand will need to reposition mid-air to open the lid while the other hand adjusts for single-handed grasping. The policy would attempt different actions, most of which would drop the box and terminate the episode.

This motivates us to propose our curriculum approach, to let the policy explore different strategies in a less fragile setting. Our core idea is using \textit{virtual object controllers}: they drive the object to follow the targets on its own, such that the policy can learn through the entire sequence and be discouraged from myopic strategies.

% where the policy must explore a high-dimensional action space and maintain stable object control.
% For instance, a policy must anticipate that one hand will need to reposition mid-air to open a box lid after both hands have lifted it. RL training often fails in such settings due to compounding errors from suboptimal early strategies.
% This motivates us to propose a curriculum learning approach.
% The core idea is to first \textit{soften} the otherwise brittle task, by using what we call ``virtual object controllers" that push the object on its own, such that the policy can optimize for the auxiliary rewards to gradually learn the interaction strategies without worrying about the object tracking. This reduces the exploration burden and lets the policy gradually take over manipulation.

% allowing the policy to learn in a less fragile setting. \ajay{weird language imo -- maybe just say the idea is to allow the agent to focus on different reward terms like imitation, and then contact, gradually without worrying about the object tracking, to reduce the exploration burden and get the agent in the ballpark}

\textbf{Virtual Object Controllers.} 
We treat the demonstration states $G$ as control goals and apply virtual spring-damper constraints that move the object along its target trajectory. Initially, the virtual controllers handle most of the object movement; over time, the controller's influence is gradually reduced, requiring the policy to assume greater control to complete the task. They controllers are implemented using privileged information in simulation. Each object is equipped with six virtual 1-DoF joints for its base pose and a 1-DoF joint for articulation, and all joints are actuated by PD controllers~\cite{pid}. At every timestep, these controllers apply virtual forces based on the error between the current object state and control targets from the demonstration. The control strength is parametrized by gain parameters ($k_p$, $k_v$), which are decayed over time to enable a structured hand-off to the learned policy. 

% Recall that, at every RL timestep, the interaction clip specifies one target state for the object (in our case, 6 DoF object pose and 1 DoF articulated joint). Our key idea is to use these object states as \textit{control targets}, and take advantage of simulation to apply virtual control forces directly on the object. This effectively forms a curriculum where, at the beginning, the virtual controllers move the object to achieve the dense tracking task without the policy involved. Towards the later stage of the curriculum, the object controller strengths are are reduced, such that the controllers alone cannot fully track the object and the policy must take over. 

% We leverage privileged information in physics simulation to implement these controllers: When simulating each object, we attach one 6-DoF free joint on its base link. We also attach joints for additional DoFs in articulated objects. Those virtual joints are equipped with PD controllers to track their desired motion, making the entire object actuated. The virtual joint controllers are parametrized with the same set of $kp, kv$ gains and force range $fr$. (\mandi{todo: might add softer contact parameters depending on the final method}) Using the demonstration trajectory, we compute the desired control targets for all demonstrated timesteps. 

\textbf{Curriculum scheduling.}
Algorithm~\ref{algo:curr} describes our proposed curriculum. At the beginning of curriculum training, we set high virtual controller gains with critical damping; then we exponentially decay the gains based on the policy's learning progress, which is tracked with a history of past rewards. As a result, the policy initially will consistently achieve high task reward; because it receives a weighted sum of task and auxiliary rewards, the policy learns actions that improve motion and contact rewards while avoiding disrupting the object trajectory. Later, as the object controllers weaken, the policy gradually learns to adjust its motions to maintain high task reward. Because the auxiliary rewards use a much smaller weight, the policy can deviate from the reference hand motions learned at the earlier stages in order to prioritize optimizing for high task rewards. 
\vspace{-2mm}
\section{Experiments}\label{sec:exp} 
\vspace{-2mm}
% and establishes the effectiveness of our proposed contact reward: we run RL training without curriculum and show that our reward improves performance over baselines, but all compared methods still fall short at completing the full-length demonstrations.
\paragraph{Experiment Setup.} We use hand-object data from ARCTIC \cite{fan2023arctic} (see \S\ref{app:arctic}), which includes 5 articulated objects~\cite{xu2025robopanoptes} and 7 demonstrations consisting of diverse motion sequences (picking up and reorienting objects, opening/closing lids, etc.) We evaluate our algorithm on both short- (used in prior work~\cite{Chen2024ObjectCentricDM}) and long-horizon demonstrations. We curate assets for 6 open-source dexterous robot hand models, with varying sizes and kinematic designs. We use Genesis~\cite{Genesis} for physics simulation, and PPO~\cite{ppo,rl-games2021} as the base RL algorithm. The policies share the same structure for state-based input observation spaces for all hands and tasks, and control both hands at once. See Appendix for details on RL training (\S\ref{app:rlenv}) and evaluation setup \S\ref{app:eval}.
% Our task environments will be open-sourced to facilitate future research. 
\vspace{-2mm}
\paragraph{Baseline Methods.} Due to various differences in physics simulation and training configurations, we re-implement baseline methods in our training framework and make several adaptations to ensure a fair comparison --- see \S~\ref{app:baseline} for implementation details. We compare against the following methods:
%\vspace{-2mm}
\begin{enumerate}
    \item \textbf{Kinematics Only.} Directly use kinematic retargeting~\cite{qin2023anyteleop} results as policy controller targets.
    %\vspace{-2mm}
    \item \textbf{ObjDex~\cite{Chen2024ObjectCentricDM}.} learns a high-level wrist planner for wrist base actions, and a low-level policy with task reward and the same hybrid actions as ours. We validate our re-implementation by showing improved performance on the same demonstrations used in the original results. 
    \item \textbf{Task + Auxiliary Rewards without curriculum.} To evaluate the effect of our proposed curriculum, we run RL training with only our proposed motion imitation and contact rewards. For a fair comparison, all training hyper-parameters are identical with our curriculum setting.
    \item \textbf{ManipTrans~\cite{li2025maniptrans}.} A concurrent work that fine-tunes a motion imitation model with RL using contact force rewards and a curriculum on error thresholds and physics parameters. Since the original method is evaluated on rigid objects in a different physics simulator, we re-implement their proposed curriculum while using our hybrid actions and auxiliary reward terms.
\end{enumerate}

% \vspace{-3mm}
\paragraph{Overview of Experiments.}
We present empirical results that (1) evaluate the effectiveness of \ours~against baselines and no-curriculum settings (\S~\ref{exp:baseline}); (2) ablate key components of our method (\S~\ref{exp:ablate}); (3) demonstrate \ours's applicability across various dexterous hand embodiments and utility as an evaluation framework for comparing different hand designs (\S\ref{exp:full}).
% In \S~\ref{exp:baseline}, we compare training with only task reward with training with our proposed auxiliary rewards and virtual object controller curriculum. Next, \S~\ref{exp:ablate} ablates our action output formulation and compares our object controller curriculum with other curriculum variants. In \S\ref{exp:full}, we evaluate \ours~on a larger set of dexterous hands and discuss the key findings from our empirical comparisons across different hand designs.

 \begin{figure}
     \centering
     \includegraphics[width=0.99\linewidth]{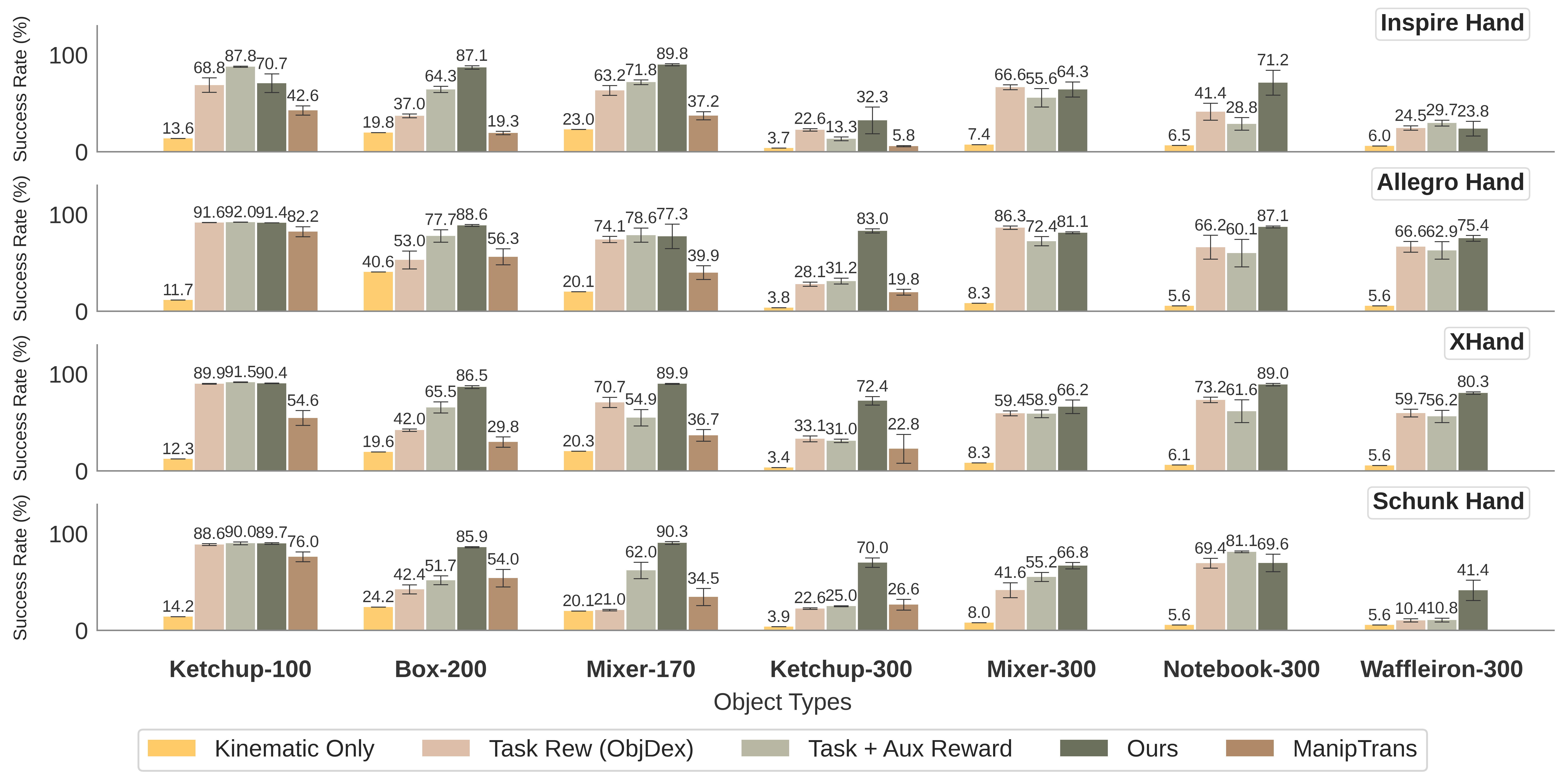}
     \caption{\small \textbf{\ours~Core Results.} We evaluate \ours~on four representative dexterous hands paired with seven demonstrations with diverse objects and motion sequences. We compare between direct replay of kinematic retargeting results (``Kinematic Only"), training with only a task reward (``Task Rew (ObjDex)", i.e., our re-implementation of ObjDex~\cite{Chen2024ObjectCentricDM}), training with both task and auxiliary rewards (``Task + Aux Reward"), and with our proposed auxiliary rewards and curriculum (``Ours"). With rare exceptions, \ours~demonstrates clear improvements over baseline methods, especially on long-horizon tasks with more complex motions.}
     \vspace{-3mm}
     \label{fig:main}
 \end{figure} 
\vspace{-3mm}
% \subsection{Baseline Comparisons}
\subsection{\ours~Main Results}
\label{exp:baseline}
%\vspace{-3mm}
We evaluate \ours~and baseline methods on four representative dexterous hands (Inspire~\cite{inspire}, Allegro~\cite{allegro}, Xhand~\cite{xhand}, and Schunk~\cite{schunk}) and seven demonstration clips (see \ref{app:qual} for visualization). Averaged success rates for each task are reported in Figure~\ref{fig:main}. The key takeaways are the following:
\vspace{-3mm}
\paragraph{\ours~consistently improves performance on all hands and tasks.} We highlight the rightmost four columns in Fig.~\ref{fig:main}, which correspond to long-horizon demonstrations with complex motion sequences\footnote{For instance, `Waffleiron-300' requires the policy to pick up the object, open and close the lid, flip it back and forth, then open and close the lid again, all mid-air (see Appendix \S\ref{app:qual} for a visualization)}. Task reward alone falls short on these clips; incorporating auxiliary rewards ('Task + Aux Rewards') improves performance on some tasks, but the gains are inconsistent. In contrast, \ours~significantly outperforms no-curriculum setting despite using the same rewards. 

Task reward and hybrid actions can achieve reasonable performance on short-horizon tasks: our re-implementation of ObjDex~\cite{Chen2024ObjectCentricDM} (`Task Rew (ObjDex)' in Fig.~\ref{fig:main}) performs better than their original reporting on the same demonstrations (left three columns in Fig.~\ref{fig:main}, detailed in \S\ref{app:baseline}). The kinematic retargeting results alone cannot complete the task (`Kinematics Only') --- our videos qualitatively show that they visually align well with human hands, but the actions cannot achieve more than slightly lifting up each object. 

% overall tasks and hands, ObjDex Avg: 0.54, add aux rew Avg: 0.58, OURS Avg: 0.75 
% group by first 3 objects and last 4 clips
% Method: ObjDex
% Short Clips: Avg: 0.62 +/- 0.06
% Long Clips: Avg: 0.48 +/- 0.06
% Method: Task + Aux
% Short Clips: Avg: 0.74 +/- 0.04
% Long Clips: Avg: 0.46 +/- 0.05
% Method: OURS
% Short Clips: Avg: 0.86 +/- 0.02
% Long Clips: Avg: 0.67 +/- 0.05

\begin{figure}
    \centering
    \includegraphics[width=0.9\linewidth]{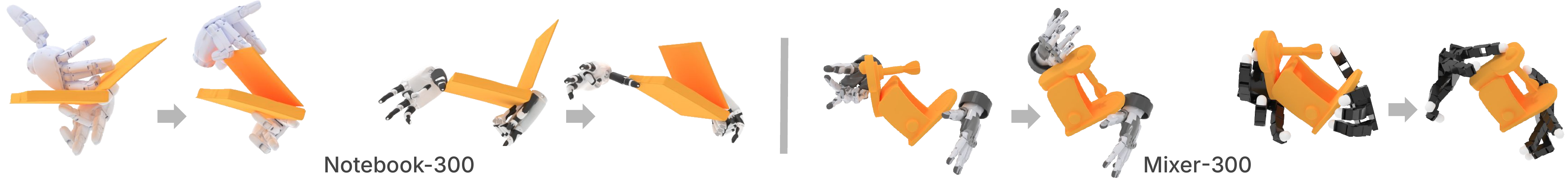}
    \caption{\textbf{\ours~Hand-Specific Strategies.} \ours~enables the policy to learn task strategies that adapt to their hardware constraints. We show snapshots of trained policy rollouts for different hands on the same task: left side shows XHand and Inspire Hand for Notebook-300 task; right side shows Schunk Hand and Allegro Hand for Mixer-300 task.  
    % On Notebook-300, XHand policy follows human demonstrator to left hand to hold up the object and \textit{right} hand to open the book cover, whereas Inspire Hand policy uses both hands to stabilize the object and only \textit{left} hand thumb to close the cover. 
    }
    \label{fig:qual}
\end{figure}

% On the shortest Ketchup-100 task, this setup alone can achieve high tracking accuracy with over $90\%$ success rate. 
%\vspace{-3mm}
% \paragraph{Auxiliary rewards show improvements on some tasks} -> fold it into below paragraph
%\vspace{-3mm}
\paragraph{\ours~lets the policy learn task strategies that adapt to hardware constraints.} The auxiliary rewards do not always align with the best task strategy, but instead act as soft guidance to serve the curriculum, giving the policy flexibility to explore. Qualitatively, we observe that the policies may deviate from the motion and contact guidance and learn different strategies: as shown in Fig.~\ref{fig:qual}: on Notebook-300, the XHand policy follows the human demonstrator to use the left hand to hold up the object and the right hand to close the cover; however, for the smaller, less-actuated Inspire Hand, the policy learns to use both hands to stabilize the object and close the cover. On Mixer-300, the Allegro Hand fingers are long enough to close the lid easily, but the Schunk Hand policy shows more wrist movements to achieve the same effect.

\subsection{\ours~Ablations}
\label{exp:ablate} 

\textbf{Action Ablations.} We compare our hybrid action formulation with: (1) absolute actions on all joints and (2) less-constrained residual actions on wrist joints, in which the wrist joint limits are set to cover the maximum motion range in the entire demonstration clip. We train in the no-curriculum setting, use a subset of tasks and hands and average over three seeds for each method. Results are shown in Fig.~\ref{fig:action-ablate}. While all methods benefit from using auxiliary rewards, using more restrictive bounds on wrist motion results in the best overall performance.

\textbf{Curriculum Ablations.} In Fig.~\ref{fig:main}, we compare \ours~with ManipTrans~\cite{li2025maniptrans}, which uses a curriculum over error thresholds for motion and object poses plus gravity and friction parameters. We observe that it achieves no clear improvements over the no-curriculum setting, and training is less stable: given the same budget of RL iterations, the ManipTrans policy initially achieves high task reward, but performance drops as the curriculum progresses and cannot recover. This indicates that merely decaying physics parameters is not sufficient for long-horizon tasks with articulated objects, which needs a stronger guidance to completely solve the task until the policy gradually takes over. 

\vspace{-2mm}
\subsection{Hand Embodiment Analysis}
\label{exp:full}
\begin{wrapfigure}{r}{0.45\linewidth} 
 %\vspace{-5mm}
 \centering
\includegraphics[width=0.99\linewidth]{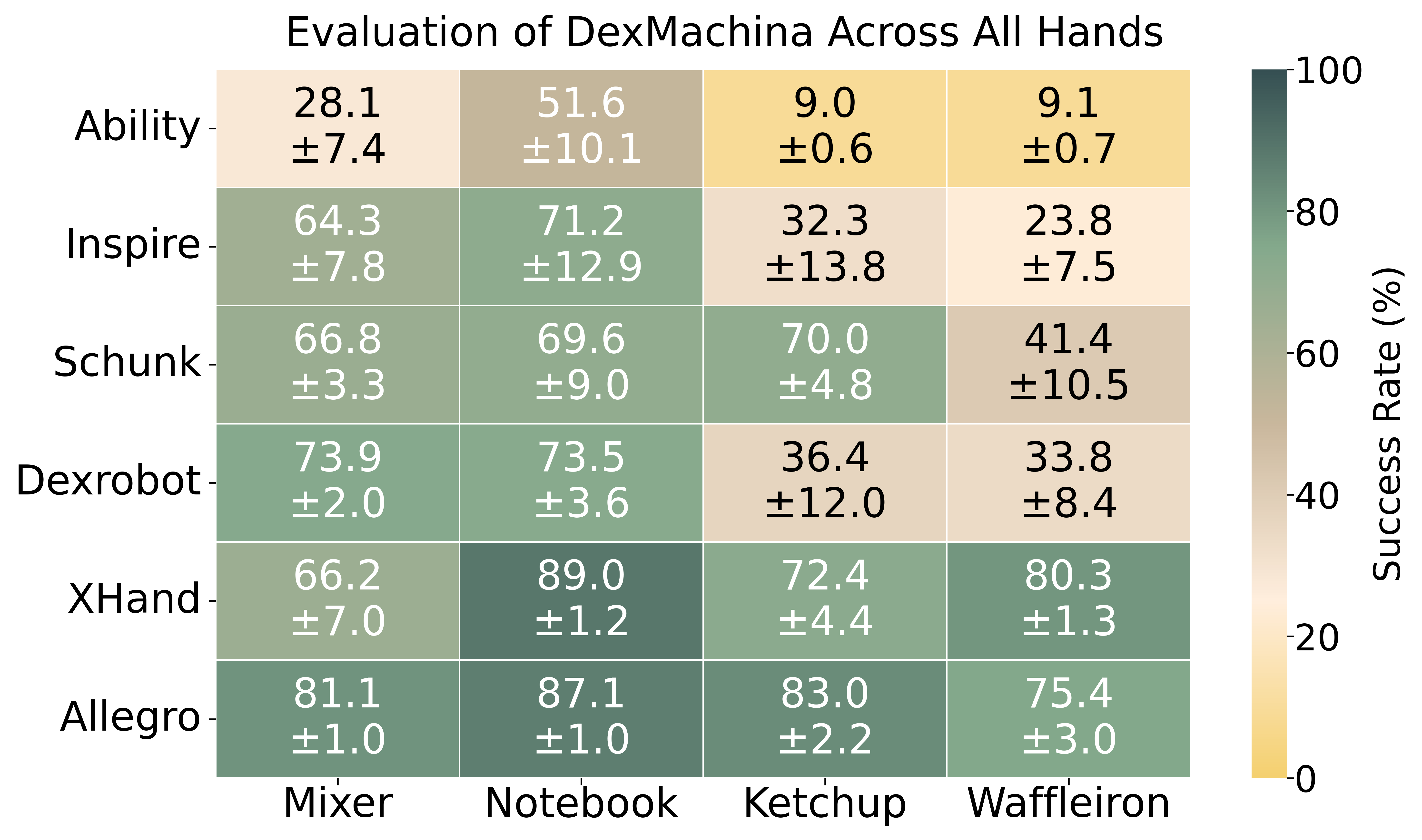}
  %\vspace{-3mm}    
  \caption{\small Full evaluation of all six hands using \ours~on long-horizon tasks.}
    \label{fig:heatmap} 
     \vspace{-15mm}
\end{wrapfigure}

\vspace{-2mm}
After validating that \ours~achieves functional retargeting across various tasks and hands, we now use our algorithm and benchmark for a functional comparison between different dexterous hands. We focus on the four long-horizon tasks from \S\ref{exp:baseline}, and evaluate \ours~on two additional hands, Ability~\cite{ability} and DexRobot Hand (see Fig.~\ref{fig:heatmap}). We discuss the following key findings:

\textbf{Larger, fully-actuated hands achieve both higher final performance and better learning efficiency.} The Allegro Hand, despite being less anthropomorphic in appearance, is surprisingly capable due to its long finger length providing stability for in-hand / in-air manipulations. \textbf{Similarity in size is less important than degrees of freedom.} For instance, the Inspire, Ability, and Schunk Hand all have similar sizes, but Schunk has actuated fingertips and a foldable palm, and achieves on average better performance than Inspire and Ability. \textbf{Although less-actuated hands are more similar to human hands in appearance, learned strategies are less human-like than the bigger but more capable hands.} Because all hands use the same set of human hand motion references (both as base wrist actions and motion rewards), the extent to which a policy deviates from human guidance is determined by their size and kinematic constraints. As a result, hands like Inspire and Ability often need different strategies to complete the task.

% \begin{itemize}
%     \item Larger, fully-actuated hands achieve both higher final performance and better learning efficiency. The Allegro Hand, despite being less anthropomorphic in appearance, is surprisingly capable due to its long finger length providing stability for in-hand / in-air manipulations. 

%     \item Similarity in size is less important than degrees of freedom. For instance, the Inspire, Ability, and Schunk Hand all have similar sizes, but Schunk has actuated fingertips and a foldable palm, and achieves on average better performance than Inspire and Ability. 
    
%     \item Although less-actuated hands are more similar to human hands in appearance, learned strategies are less human-like than the bigger but more capable hands. Because all hands use the same set of human hand motion references (both as base wrist actions and motion rewards), the extent to which a policy deviates from human guidance is determined by hardware constraints. As a result, hands like Inspire and Ability often need different strategies to complete the task.
%     % \item \mandi{todo: check results first} The more  dexterous hands 
% \end{itemize}
Naturally, our conclusions are limited by the objects and tasks that we test on: for instance, the larger hands will not perform well for smaller objects (e.g., tweezers). However, our evaluation framework can be easily extended to add new dexterous hands and test tasks or objects.

\section{Conclusion}
\label{sec:conclusion}
We present \ours, a curriculum-based RL algorithm for functional retargeting, where the key idea is to use virtual object controllers that let the policy easily explore task strategies under motion and contact guidance. In our simulation benchmark with a diverse set of tasks and dexterous hands, we show \ours~significantly outperforms baseline methods and enables functional comparison across different dexterous hand designs. We hope our algorithm and benchmark environments will provide useful platform for identifying desirable dexterous hand capabilities and lower the barrier for contributing to future research.

 % \begin{figure}
 %     \centering
 %     \includegraphics[width=0.9\linewidth]{figs/table420_pos0.05_rot0.2_arti0.2.png}
 %     \caption{\small Baseline comparisons without curriculum training. We remark that 1) our proposed contact reward improves performance over imitation-only rewards, 2) training with only task reward leads to the worst performance.  TODO: replace last 3 column with better results for green bars}
 %     \label{fig:enter-label}
 % \end{figure} 

%===============================================================================
% Limitation does NOT need to be within the page limit
\clearpage
\section{Limitations}
\label{sec:limit}

% In this work we trained state-based policies that use privileged information such as object pose -- 

\ours~has a few key limitations. 
% that point to interesting directions for future work. 
First, our policy uses state-based input that relies on privileged information from the physics simulator; this information can be challenging to acquire in the real world. This limitation can be addressed with either vision-based RL policy training~\cite{DextrAH-G}, or more practically, a distillation setup that trains visuomotor policies using demonstration data generated from state-based policies, as seen in prior work~\cite{Chen2024ObjectCentricDM}. Second, our problem formulation assumes access to high-quality human hand-object demonstration data, which requires object reconstruction and accurate pose tracking for both human hands and articulated objects. Such data can be expensive to collect and requires careful curation (e.g. ARCTIC~\cite{fan2023arctic} uses a motion-capture system with dense manual annotations and post-processing). Future work could investigate alternative methods to scale up data collection: one direction is leveraging recent progress in 3D generative models and reconstruction methods. Third, because we use open-source assets for the dexterous hands and estimate physical properties (such as mass, inertia, and collision shapes), the simulated hands might fail to capture some of the dynamics and capabilities of the real hardware. To address this, more careful tuning with real reference hardware will be needed; ideally, accurate simulation models would be provided directly by manufacturers. Lastly, our learned RL policies have not yet been evaluated in real-world settings on the examined range of dexterous hands due to lack of hardware access. With inputs from the community, we aim for our simulation benchmark to enable accessible research and evaluation of hand designs without requiring physical hardware; furthermore, our learned policy can be used as teacher policies to be distilled for sim-to-real transfer, which prior work in similar settings has demonstrated~\cite{Chen2024ObjectCentricDM,li2025maniptrans}. 

% The acknowledgments are automatically included only in the final and preprint versions of the paper.
\acknowledgments{This work was supported in part by NVIDIA, and by NSF Awards \#2143601, \#2037101, and \#2132519. The views and conclusions contained herein are those of the authors and should not be interpreted as necessarily representing the official policies, either expressed or implied, of the sponsors. The authors would like to thank current and former colleagues at NVIDIA: Kelly Guo, Milad Raksha, David Hoeller, Bingjie Tang for their help with physics simulation environments and insightful discussions during algorithm development; and all the members of REALab at Stanford University for providing useful feedback on initial drafts of the paper manuscript.}

%===============================================================================

\bibliography{refs}  % .bib 
\appendix 

\textcolor{red}{\textbf{Please see our submission website: \url{dexmachina-submission.github.io} }}
\section{Demonstration Data Processing Details}
\label{app:arctic}
\subsection{ARCTIC Demonstration Selection and Curation}
We use a subset of hand-object interaction clips from the ARCTIC dataset~\cite{fan2023arctic}, which contains articulated object scans and interaction sequences with tracked MANO~\cite{MANO:SIGGRAPHASIA:2017} hand poses and object states. Each selected clip is defined by an object (e.g. `box'), a subject tag (e.g. `s01-u01') identifying the human demonstrator, and a (start, end) tuple to trim the sequence to a fixed length, hence the number of used frames $T$ is defined as $T=(\text{end} -\text{start})$

\paragraph{Dexterous Hand Asset Processing.} All of the dexterous hands in our experiments are curated from open-source URDF models and manually edited to add 6-DoF wrist joints that achieve a `floating hand' style wrist actuation. Some of the dexterous hand models require additional processing for stable simulation, such as manually changing mass or inertia values, running convex-decomposition to improve collision mesh quality, and adding dummy links to fingertips to record and track keypoint positions. For each dexterous hand, we manually specify which finger links should match with which MANO~\cite{MANO:SIGGRAPHASIA:2017} hand joints(e.g. thumb to human thumb), which is required by the kinematic retargeting~\cite{opentelev} algorithm. The kinematic retargeting results are also used for controller gain tuning, which ensures the dexterous hand controllers are stable and fast enough to match the desired human hand movement and speed within reasonable error. 

\subsection{Object-aware Retargeting Post-processing}\label{app:para-ret}
 \begin{wrapfigure}{r}{0.5\linewidth} 
 \centering
 \includegraphics[width=0.99\linewidth]{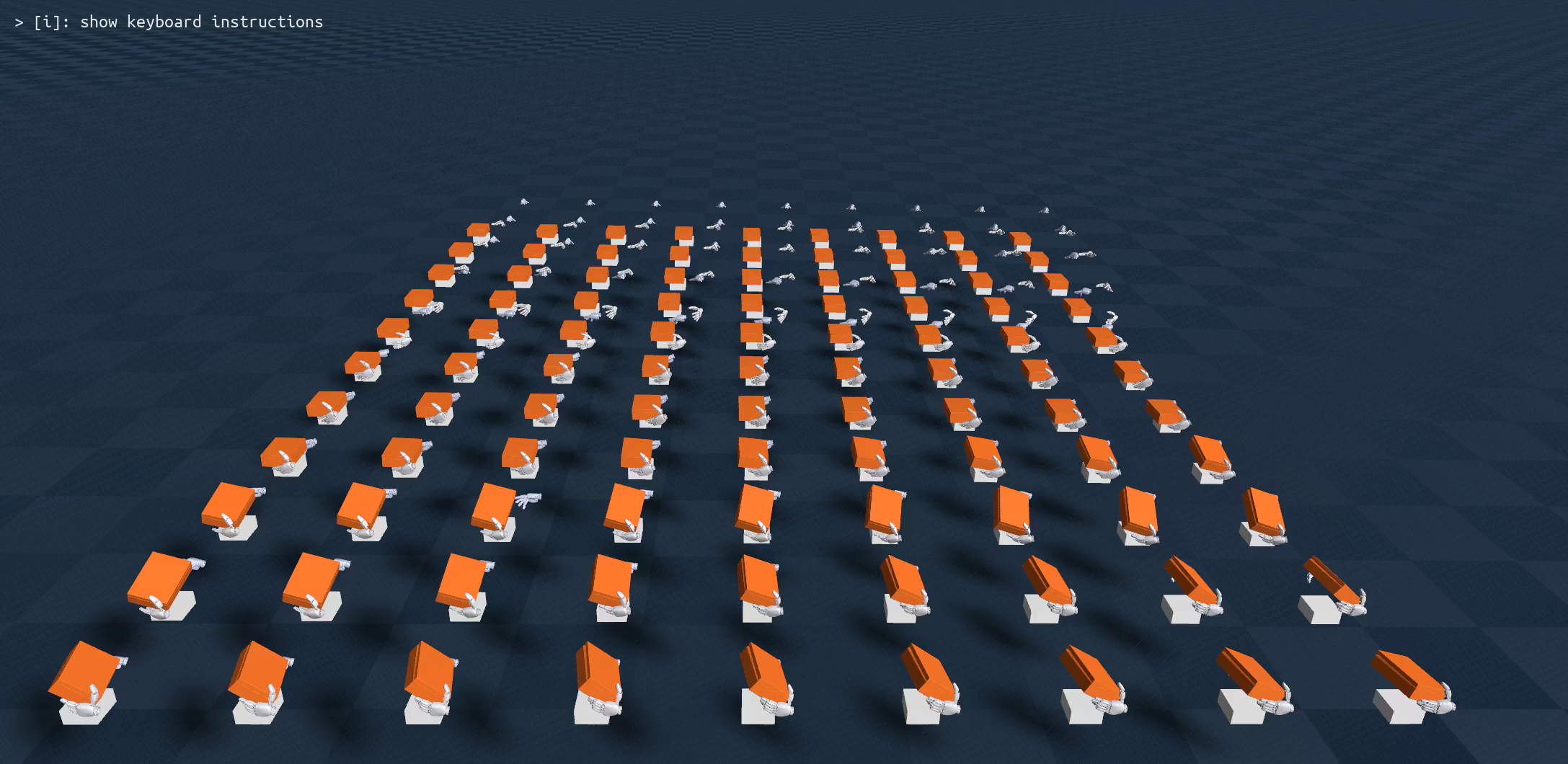}
    \caption{We perform an improved retargeting scheme over pure kinematic retargeting}
    \label{fig:obj-retarget} 
\end{wrapfigure}
Because we use densely-tracked human hand and object interaction as demonstration, a purely kinematic retargeting algorithm~\cite{opentelev} on fingertip positions results in frequent penetration with the object, which leads to damaging base-actions during policy learning, and infeasible keypoint positions which we use for imitation reward computation. To address this, we run the simulation for each pair of dexterous hands, and for each demonstrated timestep, we fixate the object to its target state (both root pose and object joint angle), and set retargeted joint values as control targets. This process lets the simulation to resolve collision and 

Then we record the achieved joint values and keypoints to use for policy learning. In implementation, this process can be easily parallelized in simulation, which we illustrate in Figure ~\ref{fig:obj-retarget}. 

\subsection{Hybrid Action Outputs.}\label{app:hybrid}

Formally, we use the following notations:
\begin{itemize}
  \item $\text{clip}(x, a, b)$: elementwise clamp of input value $x$ between $a$ and $b$
  \item $a_t \in \mathbb{R}^J$: the policy's joint action output at time $t$ clipped to $[-1, 1]$, i.e. $a_t=\text{clip}(\pi_{\theta}(o_t), -1, 1)$
  \item $q_t^{(i)}$: the target position for the $i$-th joint at time $t$ 
  \item $\mathcal{I}_f \subset \{1, \dots, J\}$: indices corresponding to the finger DOFs
  \item $\mathcal{I}_w^{\text{T}}\subset \{1, \dots, J\}$: indices of the three wrist translation DOFs, $|\mathcal{I}_w^{\text{T}}| = 3$
  \item $\mathcal{I}_w^{\text{R}}\subset \{1, \dots, J\}$: indices of the three wrist rotation DOFs, $|\mathcal{I}_w^{\text{R}}| = 3$
  \item $\mathbf{q}_t \in \mathbb{R}^J$: the \textbf{retargeted} joint values at time $t$
  \item $s_{\text{T}}, s_{\text{R}}$: scaling factors for translation and rotation actions respectively
  \item $\ell, u \in \mathbb{R}^J$: vectors of lower and upper joint limits
  \item $\hat{q_t}\in \mathbb{R}^J$: the joint target values sent to the policy’s controller 
\end{itemize}

Then, the joint target computation is defined as:
\begin{align*}
a_t^{\text{wrist-T}} &= a_t[\mathcal{I}_w^{\text{T}}] \in \mathbb{R}^3, \quad
q_t^{\text{wrist-T}} = \mathbf{q}_t[\mathcal{I}_w^{\text{T}}] + s_{\text{T}} \cdot a_t^{\text{wrist-T}} \\
a_t^{\text{wrist-R}} &= a_t[\mathcal{I}_w^{\text{R}}] \in \mathbb{R}^3, \quad
q_t^{\text{wrist-R}} = \mathbf{q}_t[\mathcal{I}_w^{\text{R}}] + s_{\text{R}} \cdot a_t^{\text{wrist-R}} \\
a_t^{\text{fingers}} &= a_t[\mathcal{I}_f], \quad
q_t^{\text{fingers}} = \ell_{\mathcal{I}_f} + \frac{u[\mathcal{I}_f] - \ell[\mathcal{I}_f]}{2} \cdot (a_t^{\text{fingers}} + 1) \\
\hat{q}_t &= \text{concat}(q_t^{\text{wrist-T}},\ q_t^{\text{wrist-R}},\ q_t^{\text{fingers}})
\end{align*}

\subsection{Contact Approximation}\label{app:contact}
% For each object's rigid parts and each MANO finger link, we take all vertices on the object part mesh that have the $L_2$ distances to their nearest-neighbor on the finger link mesh below a given threshold, and average their positions into one approximate contact position. If no vertex is below the distance threshold then the contact is invalid. This results in a set of shape $(T, M, N, 3)$ 3D positions and a shape $(T, M, N)$ valid mask for each hand. 
Let: $V_o = \{v_i^o\}_{i=1}^{N_o}$ be the vertices of one object part mesh, $V_h = \{v_j^h\}_{j=1}^{N_h}$ be the vertices of one MANO hand mesh, $\gamma$ be the contact distance threshold, $N_c$ be the maximum number of raw contact approximations (we use $\gamma=0.01, N_c=50$), and $K$ be the number of collision links on a dexterous robot hand. 

First, we do contact approximation by finding object mesh vertices that, their $L_2$ distance to the nearest neighbor on the MANO mesh is within $\gamma$: for each $v_i^o$, we get $
v_j^* = \arg\min_j \|v_i^o - v_j^h\|_2$, and mark $v_i^o$ as an approximate contact point if $\|v_i^o - v_{j^*}^h\|_2 < \gamma$. When there's more than $N_c$ vertices within this threshold, we use farthest sub-sampling to get the final set of $N_c$ contacts, denoted as \( C = \{v_k^c\}_{k=1}^{N_c} \subset V_o \). 

Next, we ``retarget" the raw approximate contacts to the dexterous robot hand: let  $L = \{\ell_m\}_{m=1}^{K}$ be the center positions of the dexterous hand links, for each contact point \( v_k^c \in C \), assign it to the nearest link:
$m^* = \arg\min_m \|v_k^c - \ell_m\|_2$
For each link $\ell_m$, compute the average position of the assigned contacts:
$\bar{v}_m = \frac{1}{|C_m|} \sum_{v_k^c \in C_m} v_k^c$
where \( C_m \subset C \) is the subset of contacts assigned to link \( \ell_m \). If \( |C_m| = 0 \), then \( \bar{v}_m \) is marked as invalid.

The final outputs are:
\begin{itemize}
    \item A contact tensor \( \mathcal{C} \in \mathbb{R}^{T \times N \times K \times 3} \)
    \item A validity mask \( \mathcal{M} \in \{0,1\}^{T \times N \times K} \)
\end{itemize}
where $T$ is the number of time-steps in the demonstration clip, $N$ is the number of object parts ($N=2$ for all our articulated object assets). The exact same procedure is repeated for each dexterous hand, hence each bi-manual RL task environment has two copies of contact information with the same shapes. 

% \begin{algorithm}[h!]
% \caption{Contact Assignment from Human to Robot Hand}
% \begin{algorithmic}[1]
% \REQUIRE Object mesh vertices $V_o$, MANO hand vertices $V_h$, dex hand link centers $L = \{\ell_m\}_{m=1}^{L}$, threshold $\gamma$
% \ENSURE Contact tensor $\mathcal{C}$ and valid mask $\mathcal{M}$
% \STATE Initialize contact set $C \gets \emptyset$
% \FOR{each vertex $v_i^o \in V_o$}
%     \STATE Find nearest $v_j^h \in V_h$ by L2 distance
%     \IF{$\|v_i^o - v_j^h\|_2 < \gamma$}
%         \STATE Add $v_i^o$ to contact set $C$
%     \ENDIF
% \ENDFOR
% \FOR{each contact $v_k^c \in C$}
%     \STATE Find nearest dex hand link $\ell_{m^*}$ by L2 distance
%     \STATE Assign $v_k^c$ to link $m^*$
% \ENDFOR
% \FOR{each dex hand link $\ell_m$}
%     \STATE Collect all contacts $C_m$ assigned to $\ell_m$
%     \IF{$|C_m| > 0$}
%         \STATE Compute average contact $\bar{v}_m = \frac{1}{|C_m|} \sum_{v_k^c \in C_m} v_k^c$
%         \STATE Set $\mathcal{M}[m] \gets 1$
%         \STATE Set $\mathcal{C}[m] \gets \bar{v}_m$
%     \ELSE
%         \STATE Set $\mathcal{M}[m] \gets 0$
%         \STATE Set $\mathcal{C}[m] \gets \text{invalid}$
%     \ENDIF
% \ENDFOR
% \end{algorithmic}
% \end{algorithm}
\begin{figure}
    \centering
    \includegraphics[width=0.99\linewidth]{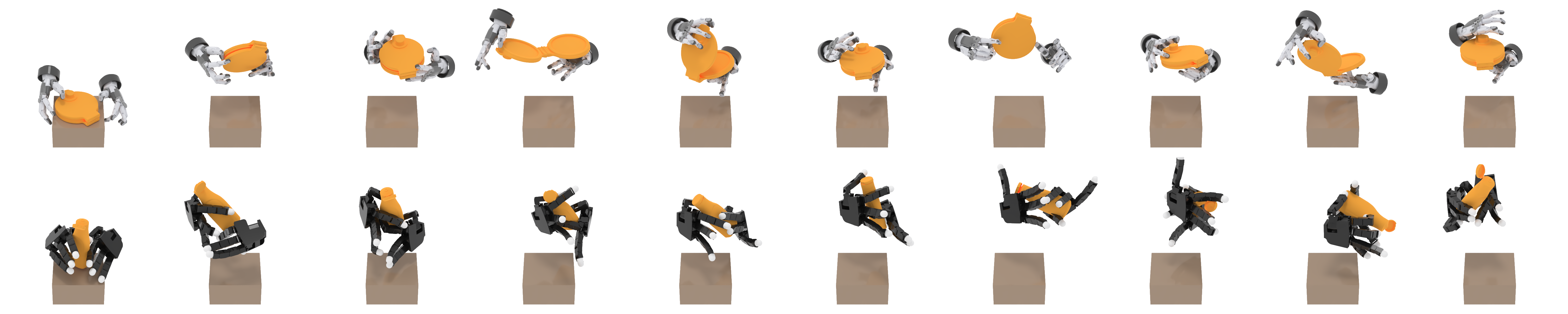}
    \caption{\small Visualization of the long-horizon tasks achieved by our trained RL policy. The brown box is used as platform surface, which follows the original ARCTIC data collection setup where objects are placed on a square cardboard box on a table surface~\cite{fan2023arctic}. }
    \label{fig:app-qual}
\end{figure}

\section{Experiment Details}
\subsection{RL Training and Evaluation Details}\label{app:rlenv}
We use Genesis for physics simulation \cite{Genesis} and PPO as the base RL algorithm implemented by the rl-games~\cite{rl-games2021} package. In the reported results for both ours and baseline methods, we use $12,000$ parallel environments for RL training on all the dexterous hands, except for Dex Hand which uses $10,000$ environments due to memory constraints. Each training run occupies either one single NVIDIA L40s or H100 GPU, and we run 5 random seeds for each demonstration and each pair of dexterous hands for all compared methods, except for action ablation experiments in \S\ref{exp:ablate} which use 3 random seeds.

\subsection{RL Policy Observation and Action Space}
We use state-based input for policy observation space: this include object states, joint targets, finger-to-object distances, and normalized hand-object contact forces.

\subsection{Details on Baseline Reimplementation}\label{app:baseline}
Our most relevant baseline methods \cite{Chen2024ObjectCentricDM, li2025maniptrans} are built with Isaac-Gym~\cite{isaacgym} with various simulation-specific implementation details.
% and missing key information on experiment setups.
To ensure a fair comparison, we have dedicated effort to ensure a faithful reimplementation using our training framework and RL environments. Some of our modifications can result in better performance for the baselines than their original reports: for example, Genesis~\cite{Genesis} uses a more stable simulation contact modeling and is more memory efficient, which enables training up to $12,000$ parallel environments with much higher learning efficiency than the Isaac Gym environments used by baselines (i.e. $2048$ for ObjDex~\cite{Chen2024ObjectCentricDM} and $4096$ environments for ManipTrans~\cite{li2025maniptrans}). We describe further details our reimplementation for each baseline below:

\paragraph{ObjDex~\cite{Chen2024ObjectCentricDM} Reimplementation Details.}
% The exact demonstration clip used for training ObjDex~\cite{Chen2024ObjectCentricDM} is not recorded. We have contacted the original authors to obtain a good estimate for the exact frame start- and end- parameters for the ARCTIC clips. The original implementation also uses a frame interpolation parameter that does not get recorded: the interpolation effectively extends the episode length for RL training to be longer than the original demonstration (e.g. an ARCTIC clip with $T$ timesteps requires training RL on $4T$ episode steps). 
To ensure a fair comparison, we have contacted the original authors to obtain their setup details that were not available publicly, including: 1. A good estimate for the exact frame start- and end- parameters for the ARCTIC clips used for training; 2. A frame interpolation multiplier that effectively extends the episode length for RL training to be longer than the original demonstration (e.g. an ARCTIC clip with $T$ timesteps requires training RL on $4T$ or $7T$ episode steps). We reuse their clip range but choose not to use the interpolation after empirically finding it to increase training time without improving task performance.

% Our reimplementation can achieve better performance than the original results. For example, on the Ketchup-30-130 clip, our reimplementation achieves consistently more than $90\%$ success rate across all tested hands, whereas the original paper reported $42\%$ success rate). 

Moreover, the original ObjDex~\cite{Chen2024ObjectCentricDM} method uses a two-level framework, where a high-level wrist planner is first learned across all ARCTIC demonstration clips, and a low-level RL policy outputs wrist residual actions. We instead directly use the kinematic retargeting results for the wrist base actions. The high-level wrist planner design assumes access to a bigger dataset and makes the low-level RL policy sensitive to the learned planner outputs --- \textbf{we hypothesize this is the main reason for why our reimplementation can achieve better performance than the original results} (e.g. On Ketchup-100, our re-implementation achieves $>90\%$ success rate for all hands, whereas the original paper reports $41.2\%$; on Mixer-170, ours achieves $>70\%$ success rate for three out of four hands, whereas the paper reports $57.6\%$).

\paragraph{ManipTrans~\cite{li2025maniptrans} Reimplementation Details.}
Because the original method did not directly evaluate on ARCTIC demonstrations (albeit the paper's appendix Section A.1~\cite{li2025maniptrans} reported \textit{qualitative} results for some ARCTIC objects), we reimplement their proposed curriculum while keeping everything else aligned with our best-performing setup (this includes hybrid action formulation, training with both task and auxiliary rewards and the same RL hyper-parameters, etc.). We follow the original method~\cite{li2025maniptrans} to decay four parameters during training, namely the thresholds for object pose errors and hand keypoint error, the z-axis gravity value, and the friction parameter, which we write as $\epsilon_{\text{object}}^P, \epsilon_{\text{object}}^R, \epsilon_{\text{finger}}, g_{\text{gravity}}, \mu$, respectively. We modified Genesis~\cite{Genesis}'s rigid solver to support modifying the gravity vector during RL training. ManipTrans~\cite{li2025maniptrans} does not disclose the exact decaying schedule for these parameters or the range for gravity and friction parameters, hence we choose the same exponential scheduler to stay consistent with our virtual object controller curriculum, and choose a range of $g_{\text{gravity}}\in[0, -9.81], \mu\in [4.0, 1.0]$. More specifically, given a max iteration $I$, desired range of parameters, and a decay interval $v$, the parameter is decayed every $v$ iterations; after the parameters reach their final values, training proceeds for another fixed number of iterations (this is also aligned with our method). The exponential schedule depends on the given max iterations $\mathcal{I}$, for each parameter $\omega \in\{ \epsilon_{\text{object}}^R, \epsilon_{\text{finger}}, g_{\text{gravity}}, \mu\}$: its value at a given training iteration can be written as
$\omega_{\text{current}} = \omega_{\text{init}} \cdot \left(\frac{\omega_{\text{final}}}{\omega_{\text{init}}}\right)^{t/I}$. Note that we use a pseudo value $\bar{g}_{\text{gravity}}\in[9.81, 0]$ because the decay computation assumes positive bounds, and the actual applied gravity is $g_{\text{gravity}} = 9.81 - \bar{g}_{\text{gravity}}$.

\subsection{Policy Evaluation Setup}\label{app:eval}

\paragraph{Evaluation Across Random Seeds.} For each method and task, we run 5 random seeds; each seed run saves a best policy checkpoint based on cumulative task reward, and each checkpoint is evaluated for 20 episodes. For each evaluation episode, we record the achieved object states (both pose and revolute joint angle) and compare against the demonstration trajectory. 

\paragraph{Performance Metrics.}
Our functional retargeting task requires a manipulation policy to achieve articulated object tracking, which involves balancing both pose and joint angle errors over long time sequences. For performance reporting, prior work has explored per-step success rate~\cite{Chen2024ObjectCentricDM} or tracking error~\cite{li2025maniptrans}, but both have clear shortcomings: success rate reporting is based on how many timesteps out of the entire demonstration a policy can move the object to track within given error thresholds, hence the results are highly sensitive to the threshold values (this case requires 3 different thresholds for position, rotation, and joint angle errors), which also depend on the object size and geometries. Reporting tracking errors can be accurate, but it shows three different errors for every task, hence making it difficult to derive high-level comparisons and takeaways from experiment results. To address these limitations, we propose to follow prior work in object pose tracking~\cite{foundationposewen2024,posecnn,clutter} and use a similar ADD-AUC\footnote{ADD stands for Average Distance, AUC stands for Area Under Curve, we don't use ADD-S because we have the exact matching targets from the demonstration} metric with the key difference that we compute ADD for each object part separately (to accommodate articulated objects) and average ADD results before computing AUC. We found this to be a less-sensitive metric that still reports one success rate value for each method while reflective of the qualitative results from policy roll-outs.

\section{Additional Experiment Results}
We visualize keyframes of our long-horizon manipulation tasks in Fig.~\label{app:qual}. Please see supplementary videos for additional qualitative results for policy roll-outs. The figure in the next page shows results for our action ablation experiments described in \S \ref{exp:ablate}. 

 \begin{figure} 
 \centering
\includegraphics[width=0.7\linewidth]{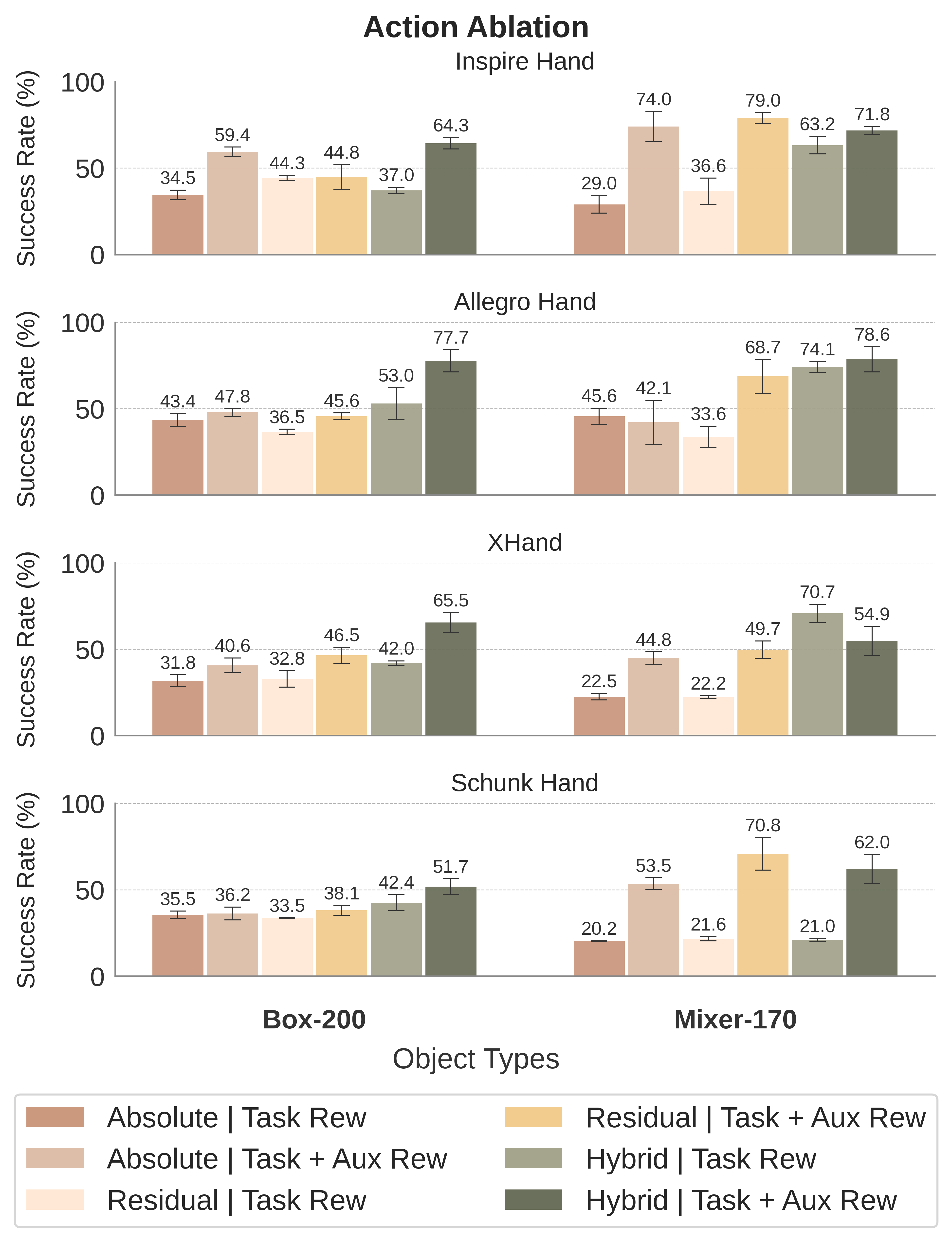}
  %\vspace{-3mm}
    \caption{\textbf{Hand Action Ablation.} We ablate on action output formulations on a subset of dexterous hands and objects and trained \textit{without} curriculum. Hybrid actions with more restrictive bounds (light and dark green bars) shows better learning performance than absolute actions and full residual actions with less wrist constraints, both in training with task rewards or with both task plus auxiliary rewards settings}
    \label{fig:action-ablate} 
     %\vspace{-5mm}
\end{figure}

%  \begin{wrapfigure}{r}{0.4\linewidth} 
%  %\vspace{-5mm}
%  \centering
% \includegraphics[width=0.99\linewidth]{figs/action_ablation_ADD_AUC.png}
%   %\vspace{-3mm}
%     \caption{\small \textbf{Hand Action Ablation.} We ablate on action output formulations on a subset of dexterous hands and objects and trained \textit{without} curriculum. Hybrid actions with more restrictive bounds (light and dark green bars) shows better learning performance than absolute actions and full residual actions with less wrist constraints, both in training with task rewards or with both task plus auxiliary rewards settings}
%     \label{fig:action-ablate} 
%      %\vspace{-5mm}
% \end{wrapfigure}

\end{document}